\pdfoutput=1

\documentclass[11pt]{article}

\usepackage[review]{ACL2023}

\usepackage{times}
\usepackage{latexsym}
\usepackage{amsmath,amssymb}
\usepackage[T1]{fontenc}

\usepackage[utf8]{inputenc}

\usepackage{graphicx}
\usepackage{cleveref}
\usepackage{subcaption}
\usepackage{microtype}

\usepackage{inconsolata}

\usepackage{multirow}
\usepackage{booktabs}
\usepackage[
singlelinecheck=false 
]{caption}
\title{\textit{WHAT}, \textit{WHEN}, and \textit{HOW} to Ground: Designing User Persona-Aware Conversational Agents for Engaging Dialogue}

\author{First Author \\
  Affiliation / Address line 1 \\
  Affiliation / Address line 2 \\
  Affiliation / Address line 3 \\
  \texttt{email@domain} \\\And
  Second Author \\
  Affiliation / Address line 1 \\
  Affiliation / Address line 2 \\
  Affiliation / Address line 3 \\
  \texttt{email@domain} \\}

\author{
Deuksin Kwon$^{1,2}$~~~
Sunwoo Lee$^1$~~~
Ki Hyun Kim$^1$
Seojin Lee$^1$~~~
Taeyoon Kim$^{1}$\thanks{equal contribution}~~~
Eric Davis$^{1}$\footnotemark[1]~~~
\smallskip 
\\
$^1$ SK Telecom, South Korea, $^2$ University of Southern California \\
\{ds.kwon, sunwoo.lois, kimkihyun, skt.kaylee, tae.y.kim, eric.davis\}@sk.com\\
deuksin.kwon@usc.edu\\
}

\begin{document}
\nolinenumbers
{\makeatletter\acl@finalcopytrue
  \maketitle
}
\begin{abstract}
This paper presents a method for building a personalized open-domain dialogue system to address the WWH (WHAT, WHEN, and HOW) problem for natural response generation in a commercial setting, where personalized dialogue responses are heavily interleaved with casual response turns. The proposed approach involves weighted dataset blending, negative persona information augmentation methods, and the design of personalized conversation datasets to address the challenges of WWH in personalized, open-domain dialogue systems. Our work effectively balances dialogue fluency and tendency to ground, while also introducing a response-type label to improve the controllability and explainability of the grounded responses. The combination of these methods leads to more fluent conversations, as evidenced by subjective human evaluations as well as objective evaluations.
\end{abstract}

\section{Introduction}\label{section:introduction}

A personalized dialogue (PD) system is capable of generating user-customized responses based on long-term memory about the user's persona, leading to more trustworthy and engaging conversations \cite{ranjbartabar2021you, xu2022long}. In our study, persona attributes cover comprehensive user-related information, such as personality, behaviors, preferences, and experience.

\begin{figure}[t!]
    \centering
    \includegraphics[width=\linewidth, height=4.5cm]{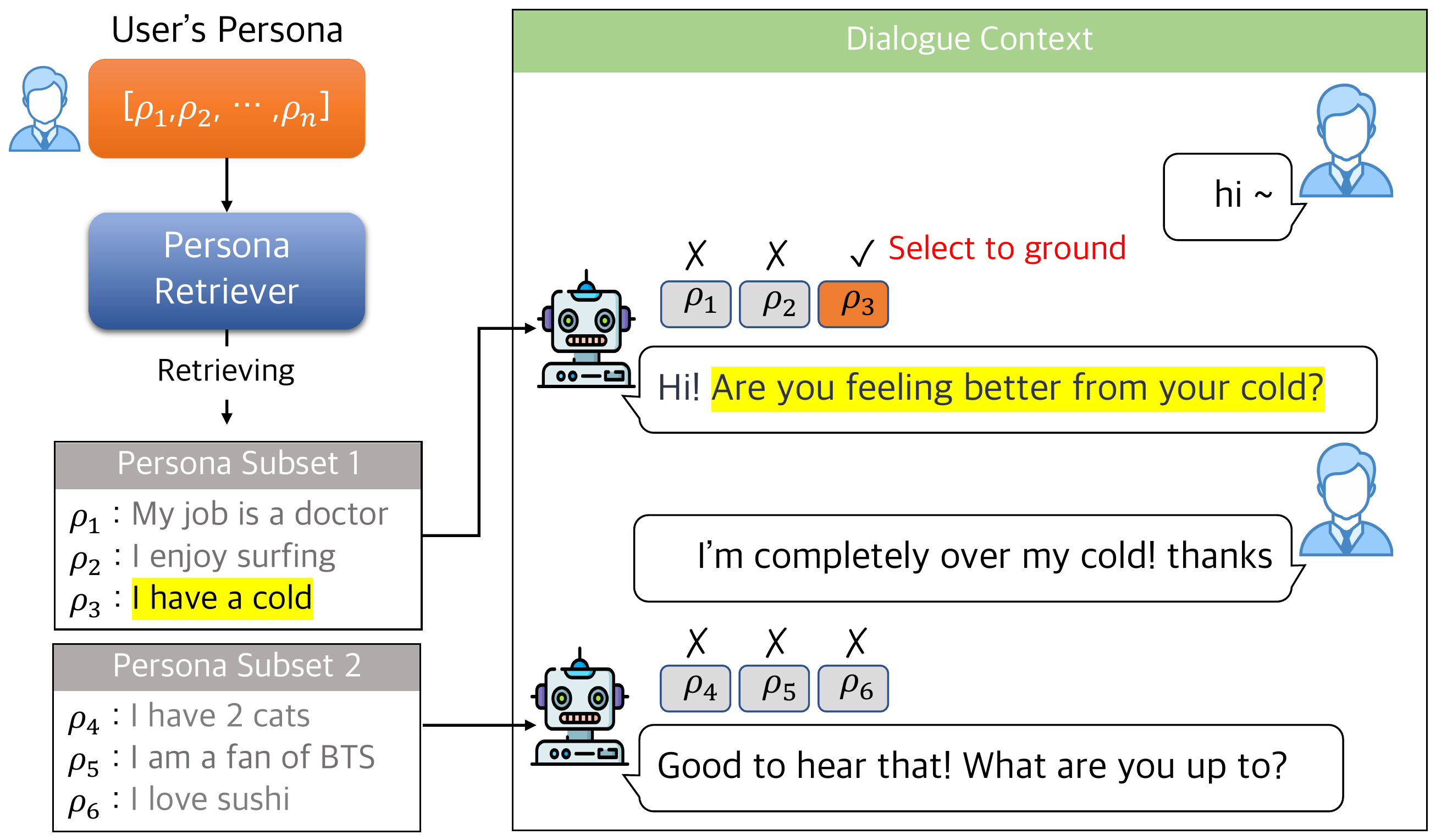}
    \caption{A sample of personalized conversation grounded on user persona. For every agent utterance, the persona attributes to be grounded in the response are retrieved by a retrieval model. Then, the agent make a decision about generating personalized response given dialogue context and retrieved persona subset.}
    \label{fig:personlized_conversatioin}
\vspace{-3ex}
\end{figure}

The key to enhanced user engagement in a PD system lies in finding a persona that is contextually relevant and appropriate, on which a model is grounded to generate a natural response. However, as shown in the example in Figure \ref{fig:personlized_conversatioin}, PD systems usually need to select relevant persona attributes from a given subset of N persona attributes, which is usually provided by external memory or retrieved from the user persona pool. Considering the fact that the agent's response is usually annotated with associated oracle persona information in the training dataset, deciding what persona attribute to select in each turn during model inference is a non-trivial problem. (We will refer to this problem as the "WHAT to ground" problem hereafter.) Another aspect to consider in a PD system is that under certain dialogue contexts, it is better not to generate a personalized response given retrieved persona attributes in order to create a more natural interaction (the second response in Figure \ref{fig:personlized_conversatioin}). As shown in the example, it is usually difficult for the retrieval module to determine whether to use persona information for response generation. We also need a model to decide when to ground persona information with a given persona subset for every turn (We will call this the "WHEN to ground" problem hereafter).

Given such a challenge, designing a user-based persona-aware PD system capable of generating engaging and human-like personalized responses requires addressing the "WHAT," "WHEN," and "HOW" (WWH) questions: 1) What personal information should be grounded given the conversation context, 2) When to generate responses using personal information, and 3) How to make natural and human-like personalized response.

\begin{figure*}[ht]
    \centering
    \includegraphics[width=\linewidth, height=6.4cm]{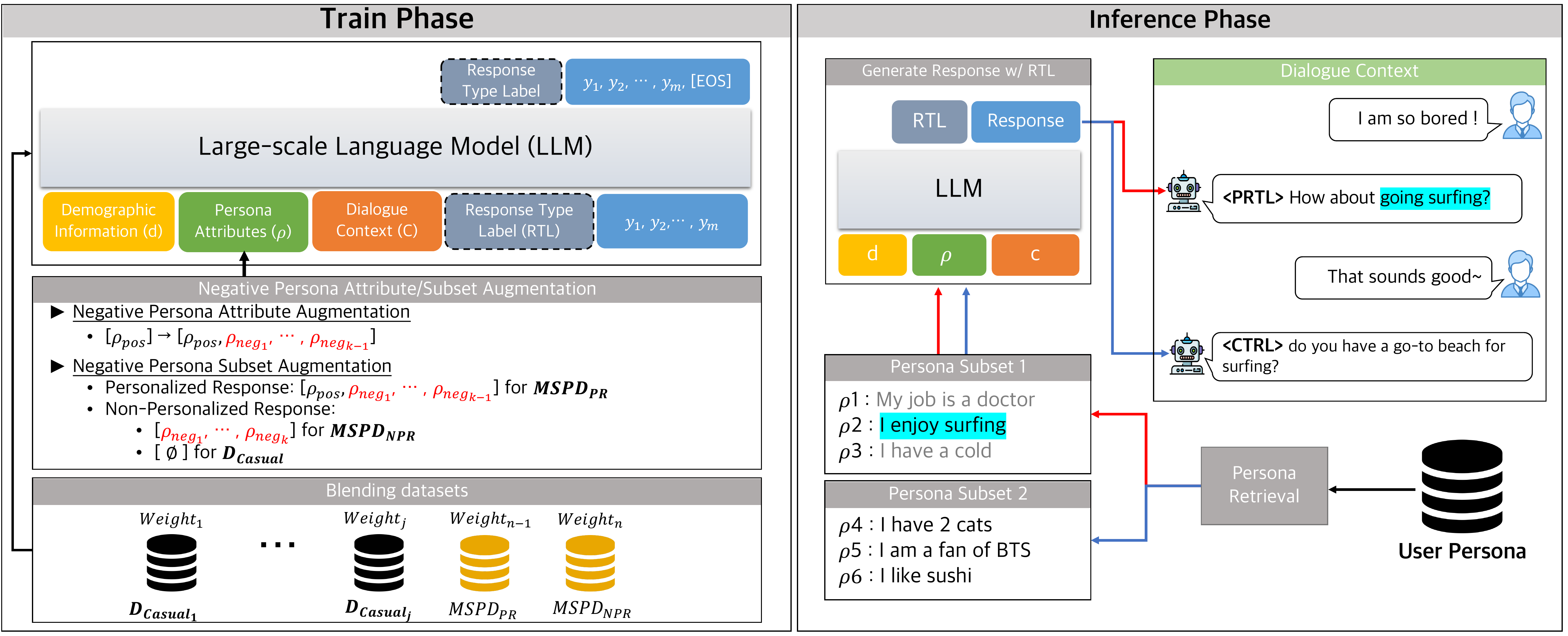}
    \vspace{-4ex}
    \caption{The overall framework of our proposed personalized dialogue system}
    \vspace{-1ex}
    \label{fig:overall_framework}
\end{figure*}

Most previous research on personalized dialogue systems has focused on generating natural responses in ideal personalized conversation settings \cite{liu2020you, dong2022know,xu2022long, fu2022there}, where issues related to heavily interleaving personalized responses with casual dialogue turns are not considered. However, we believe that these are significant problems that need to be addressed in real-world personalized conversational systems.

Large-scale Language Models (LLMs) such as GPT-3 have shown outstanding capabilities in various Natural Language Understanding (NLU) tasks and especially, in-context learning \cite{brown2020language}. However, the inherent abilities of LLMs alone are insufficient to effectively address the WWH problems in real-world service environments. Moreover, it is very tricky to generate natural and engaging personalized responses and sophisticatedly control the output of the model in multi-turn/session scenarios, relying solely on prompt engineering.


In addressing the research gap and real-world challenges, we propose a method that controls the inclination of models to generate personalized responses. Our technique blends persona-augmented datasets to construct a personalized dialogue system, thus enabling human-like natural conversations. Our approach involves the following steps:

1) We create a Multi-Session Personalized Conversation (MSPC) dataset. This trains the model to ground the provided persona information effectively for a personalized response.
2) We control the model's persona-grounding level by adjusting the blending weights of the conversational datasets. Furthermore, we enrich the dataset with negative samples of persona subsets at the turn level for model fine-tuning.
3) To enhance both generation quality and the controllability and interpretability of persona-grounded generation, we use a turn label. This label indicates whether a turn is personalized or casual and serves as one of the inputs.
Ultimately, we build a personalized dialogue system by fine-tuning an 18-billion parameter large language model (LLM). This LLM has a high level of understanding of conversation history, the ability to generate high-quality responses, and the capacity to focus effectively on given inputs, including users' personas.

We also propose four grounding type categorizations to allow for analysis of the model's grounding patterns and detailed performance in subjective evaluation using \textit{sensibleness} and \textit{specificity}, which complements the objective evaluation based on \textit{groundedness}, and \textit{fluency}.

\section{Related Work}\label{section:related_work}

\begin{figure*}[ht]
    \centering
    \includegraphics[width=\linewidth, height=6.1cm]{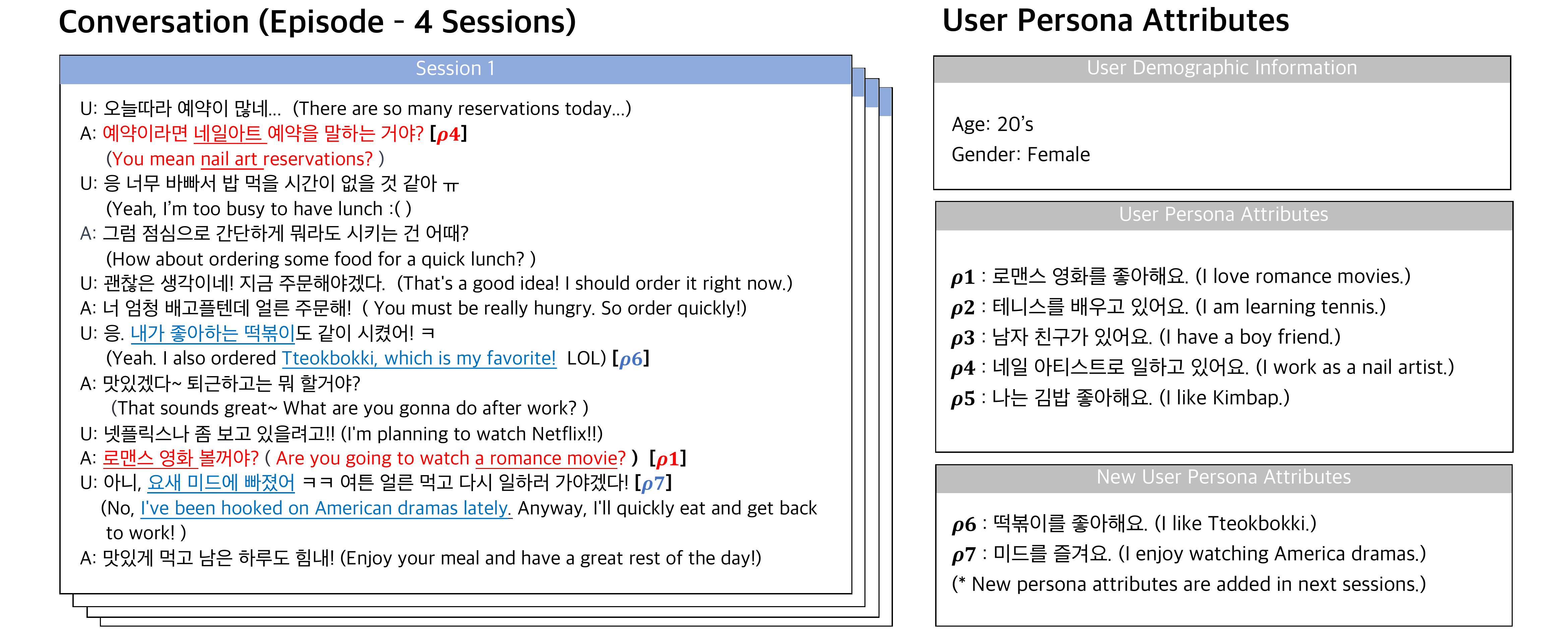}
    \vspace{-4ex}
    \caption{An example of a session from our proposed MSPD dataset. The left figure represents a dialogue between the user (U) and the agent (A), where red text and information within brackets indicate the agent's personalized responses (PR) and the index of a corresponding persona attribute. Blue text and content within brackets represent the user's new persona and its corresponding persona index. The right figure contains details about persona attributes, including the user's demographic information}
    \vspace{-1ex}
    \label{fig:example_MSPD}
\end{figure*}

Since the release of the \textbf{PersonaChat} dataset \cite{zhang2018personalizing}, methods to generate personalized responses that are consistent with or grounded on a persona have been extensively studied \cite{lee2021generating, liu2020you, xujing2021beyond, xu2022long}. Most of the studies focus on addressing the \textit{WHAT} and \textit{HOW} challenges by the use of diverse model architectures, modules, and training frameworks \cite{fu2022there, dong2022know}.


With respect to the \textit{How}, \citet{liu2020you} propose an RL-based approach for generating personalized dialogue with rewards for mutual persona perception. \citet{wu2019guiding} and \citet{fu2022there} employ variational methods to generate personalized and knowledgeable response generation. \citet{song2019exploiting} addresses both \textit{WHAT} and \textit{HOW} by generating persona-grounded responses via CVAE with a selected persona from a memory. \citet{xu2022long} and \citet{bae2022keep} also tackle the same problems via a persona retrieval module and a generator module.

 However, to the best of our knowledge, there has been no work on addressing all three \textit{WWH} questions in PD system. Therefore, considering the crucial importance of addressing the \textit{WWH} issues in a commercial system, we propose novel methods to tackle all three \textit{WWH} questions, which are mission-critical for a commercial system.

\section{Dataset}\label{section:dataset}
To develop a PD system that addresses the \textit{WWH} problems, we construct a Korean Multi-Session Personalized Dialogue dataset, which we refer to as MSPD. This dataset includes an agent that performs several unique roles, setting it apart from other PD datasets. Primarily, the agent is required to remember user persona attributes, including any persona attributes introduced during the conversation. The agent must also produce personalized responses that are both reasonably and timely grounded on the persona. The goal of this dataset is to enable a model to learn the \textit{HOW} and \textit{WHEN} of grounding.
On average, the dataset contains 4 sessions per episode, with each session consisting of 10-12 turns between the user and the agent. This format allows the agent to learn how to sustain a natural conversation flow, both within and between sessions. As illustrated in the red and blue text in Figure \ref{fig:example_MSPD}, we annotate the persona used in responses and user utterances that include personal information the model should remember. Specifically, to address the \textit{WHEN} and \textit{HOW} problems from a dataset perspective, we cap the number of personalized responses per session at two or fewer, and performed rigorous reviews to ensure the quality and appropriateness of personalized responses within conversations. This approach allows us to construct 13,469 episodes in total. Statistics and other samples from the MSPD dataset can be found in Appendices \ref{sec:appendixA} and \ref{sec:appendixB}.

Alongside the MSPD, we incorporate a variety of informal dialogue datasets, referred to as $D_{casual}$, to train a more balanced model capable of generating high-quality daily, knowledge-based, empathetic, and personalized conversations. $D_{casual}$ consists of a comprehensive collection of approximately 12.5 million utterances. We use carefully-curated Korean dialogue datasets available online\footnote{\url{https://aihub.or.kr/}}, developed by National Information Society Agency (NIA), as well as crowdsourced conversational datasets, including Korean versions of \textbf{PersonaChat}, \textbf{EmpatheticDialogues}, and \textbf{Wizard of Wikipedia} \cite{dinan2018wizard, rashkin2018towards, zhang2018personalizing}.

\section{Methodology}\label{section:methodology}
As shown in Figure \ref{fig:overall_framework}, we train our model during the training phase to address the \textit{WHAT} and \textit{WHEN} questions using a variety of methods. These include different types of negative persona augmentation, dataset blending, and response type generation.
During the inference phase, given the dialogue context and a subset of persona attributes, the model is capable of generating suitable personalized responses. These persona subsets are retrieved from individual persona attributes determined by the context of the conversation. Additionally, the model provides an explanation for its decision through response type labels (RTL). Conditioning on the RTL, allows us to explicitly control the generation of a personalized response.

\subsection{Persona-Grounded Generation}\label{section:persona_grounded_generation}

In this study, every input of the training dataset consists of user demographic information $d$ (e.g. \textit{gender}, \textit{age}), a subset of user persona $\rho^{m}$, which consists of persona attributes, and dialogue context  $c^{m} = [ u_1, a_1, u_2, a_2,\cdots, u_{m-1}, a_{m-1}, u_m]$. $u$ and $a$ refer to the user and agent, respectively, and the target response $y^{m} = [y_1^m, \cdots, y_{\ell}^m]$ is indexed to  the $m_{th}$ agent response $a_m$.





 Given the input, which is in the format of $(d, \rho^{m}, c^{m})$, we optimize the model via the conditional probability for personalized response $y^m$ and a loss function with Negative Log-Likelihood (NLL) loss that can be formulated as:
\begin{equation}
    P(y^{m}|d,p^{m},c^{m}) =\prod_{t=1}^{\ell}P(y^{m}_{t}|d,\rho^{m},c^{m}, y^{m}_{<t})\label{eq:1} 
\end{equation}
\begin{equation}
\mathcal{L}_{NLL} =-\sum_{t=1}^\ell\text{log }P(y^{m}_{t}|d,\rho^{m},c^{m}, y^{m}_{<t})
\end{equation}
where $\ell$ is the length of the target response.


\subsection{Dataset Blending}\label{section:dataset blending}

Blending a variety of conversational datasets has been shown to improve the diversity, empathy, and knowledge of a dialogue system, leading to more natural and engaging conversations \cite{smith2020blended}. By blending the MSPD, which is tailored for personalized conversations, with various types of casual dialogue datasets, $D_{casual}$, the model becomes more balanced and adept at cohesive and natural conversations. 

We define a data instance as ($c$, $r$) where $c$ and $r$ are the dialogue context and target response, respectively as described in section \ref{section:persona_grounded_generation}. We blend datasets by instance according to blending weights for each dataset.
In particular, in order to finely control the \textit{WWH} problems with the blending weights ($w$), the MSPD dataset is divided into the agent's personalized responses ($\mathcal{D}_{\text{MSPD-PR}}$) and non-personalized responses ($\mathcal{D}_{\text{MSPD-NPR}}$) (e.g., agent's red and black colored responses in Figure \ref{fig:example_MSPD}, respectively).
The final training dataset is assembled by over-sampling or under-sampling individual datasets. The training data size of individual dataset is determined by the weighted number of data instances for each dataset, defined by 

\begin{equation}
    \|\mathcal{D}_\text{i (train)}\|=
        \frac{w_i}{\sum_{j=1}^N{w_j}}\times{\|\mathcal{D}\|}
\end{equation}
where a set of $N$ conversational datasets $\mathcal{D}=\{\mathcal{D}_{casual_1},\cdots,\mathcal{D}_{casual_k},\mathcal{D}_{\text{MSPD-PR}},\mathcal{D}_{\text{MSPD-NPR}}\}$, $\mathcal{D}_i$ is $i$-th dataset in $\mathcal{D}$.


\subsection{Control of \textit{WHEN} \& \textit{WHAT} by Negative Samples}\label{section:control_of_when_what_by_negative_samples}

\textbf{Control of \textit{WHEN}} To address the \textit{WHEN} problem, it is important to control a model's propensity to generate a persona-grounded response. Given a persona, an agent must generate personalized responses at the right time to create coherent and natural conversations. Generating persona-grounded responses too frequently leads to unnatural conversations. On the other hand, a model that generates personalized responses too infrequently does not sufficiently enhance a user's engagement with the agent.

In particular situations where a persona subset is retrieved by a retrieval model at each turn, the model should generate a casual response instead of generating a personalized response, resulting in a more natural flow.
In order to learn this natural flow, we intentionally include a persona subset consisting of all contextually irrelevant persona attributes in the input for non-personalized responses. We call this a negative persona subset augmentation in our study.
This augmentation "suppresses" the model's inclination to ground too frequently. However, too much augmentation can hinder the model's ability to ground, so we perform the negative persona subset augmentation only for data in $\mathcal{D}_{\text{MSPD-NPR}}$, not all casual datasets $D_{casual}$.\\




\vspace{-1ex}
\noindent \textbf{Control of \textit{WHAT}} When a model generates a persona-grounded response, it needs to determine the \textit{WHAT}, i.e., the specific persona attribute on which to base the response. By providing both the ground-truth persona attributes, $\rho_\text{pos}$, which are relevant to the response, and "negative" persona attributes, $\rho_{\text{neg}1},...,\rho{\text{neg}_{k-1}}$, which are not relevant to the target response, the model learns to select the appropriate persona attribute(s) from multiple options given the current dialogue context. We refer to the process of adding multiple negative persona attributes to a ground truth persona as negative persona attribute augmentation.

\begin{table*}[t!]
\centering
\resizebox{\linewidth}{!}{%
\begin{tabular}{lllcccc} 
\toprule
\multirow{2}{*}{Model ID} & \multicolumn{1}{c}{\multirow{2}{*}{Method}} & \multicolumn{1}{c}{\multirow{2}{*}{Dataset}} & \multirow{2}{*}{\# Attribute} & Fluency & \multicolumn{2}{c}{Groundness} \\ 
\cmidrule[\heavyrulewidth]{5-7}
 & \multicolumn{1}{c}{} & \multicolumn{1}{c}{} &  & PPL & F1 & P-Cover \\ 
\toprule
$Model_1$ & Base (Positive Only) & $\text{MSPD}_\text{PR}$ + $\mathcal{D}_\text{casual}$ & 1 & 11.4 & 0.28 & 0.12 \\
$Model_2$ & ~ + Negative Persona Attributes Augmentation & $\text{MSPD}_\text{PR}$ + $\mathcal{D}_\text{casual}$ & 5 & 10.97 & 0.15 & 0.07 \\
$Model_3$ & ~ + Negative Persona Subset Augmentation & $\text{MSPD}_\text{PR}$ + $\text{MSPD}_\text{NPR}$ + $\mathcal{D}_\text{casual}$ & 5 & 9.37 & 0.1 & 0.05 \\
$Model_4$ & ~ + RTL Generation & $\text{MSPD}_\text{PR}$ + $\text{MSPD}_\text{NPR}$ + $\mathcal{D}_\text{casual}$ & 5 & 8.88 & 0.1 & 0.046 \\
\toprule
\end{tabular}
}
\vspace{-1ex}
\caption{The Results of Objective Evaluation}
\label{tab:objective_evaluation}
\vspace{-2ex}
\end{table*}


Finally, we vary the subset of the persona $\rho$ in \eqref{eq:1} for negative persona augmentation depending on the response type:
\vspace{-1ex}
\begin{equation*}
    \rho=\left\{
    \begin{matrix}
        \rho_{\text{npr}}\:  \text{\small{for non-personalized response}}\in\mathcal{D}_{\text{MSPD-NPR}} \\
        \rho_{\text{pr}}\:  \text{\small{for personalized response}}\in\mathcal{D}_{\text{MSPD-PR}} \\
        \rho_{c}\: \text{\small{for casual response}}\in\mathcal{D}_{\text{casual}} \\
    \end{matrix}\right.
\end{equation*}
, where $\rho_\text{npr}=\{\rho_{\text{neg}_1},\cdots,\rho_{\text{neg}_k}\}, \\ \rho_\text{pr}=\{\rho_\text{pos},\rho_{\text{neg}_1},\cdots,\rho_{\text{neg}_{k-1}}\}\text{, and }\rho_\text{c}=\phi$.





\subsection{Controllability \& Explainability via Response Type Label}\label{section:control_of_when_via_response_type_label}
\textbf{Controllability} In a commercial setting, it is often necessary to determine whether to generate a personalized response based on business logic. For instance, this might include deciding when the agent should proactively send a message to users. We can exert explicit control over the model's decision regarding the \textit{WHEN} by employing Response Type Labels (RTL), denoted as $\small\text{<RTL>}$.

First, we train the model to generate both a response and corresponding RTL token: $P(\small\text{<RTL>},y |d,\rho,c)$ in \eqref{eq:1}. We have pre-defined special tokens $\small\text{<PRTL>}$ for personalized response type labels and $\small\text{<CRTL>}$ for casual response type labels. Then, at inference time, we can insert the RTL to generate a response that corresponds to the response type: $y\sim{P_\theta(\cdot|d,\rho,c,\small\text{<PRTL>})}$ or $y\sim{P_\theta(\cdot|d,\rho,c,\small\text{<CRTL>})}$.\\

\noindent \textbf{Explainability} Error analysis is a crucial element in commercial systems for swift debugging and resolution of issues. However, this process can often be labor-intensive, typically involving a manual review of log data to evaluate the quality and appropriacy of generated personalized responses. Therefore, besides enhancing controllability, we also employ the Response Type Label (RTL) to improve the explainability of the model's generated responses. In this regard, the level of explainability provided by the RTL facilitates easier and more efficient error analysis, leading to improved service operation.


\begin{table}[t]
\centering
\resizebox{\linewidth}{!}{%
\begin{tabular}{ccccccc} 
\toprule
\multicolumn{1}{c}{\multirow{2}{*}{Model}} & \multicolumn{3}{c}{Blending Weight} & \multicolumn{3}{c}{Evaluation} \\ 
\cmidrule[\heavyrulewidth]{2-7}
\multicolumn{1}{c}{} & $\mathcal{D}_\text{casual}$ & $\text{MSPD}_\text{PR}$ & $\text{MSPD}_\text{NPR}$ & F1 & P-Cover & PPL \\ 
\toprule
\multirow{4}{*}{\begin{tabular}[c]{@{}c@{}} $Model_{3}$\\(Negative Persona Subset Aug.\\+ 5 Negative Persona Attribute Aug.)\end{tabular}} & 0.94 & 0.5 & 0.1 & 0.14 & 0.06 & 10.46 \\
 & 0.92 & 0.5 & 0.3 & 0.12 & 0.05 & 10.04 \\
 & 0.90 & 0.5~ & 0.5 & 0.11~ & 0.05 & 9.91 \\
 & 0.87 & ~0.5 & 0.8 & 0.1~ & 0.05 & 9.33 \\
\bottomrule
\end{tabular}
}
\vspace{-1ex}
\caption{Evaluations with Different Blending Weights}
\label{tab:blending_weight}
\vspace{-2ex}
\end{table}

\section{Experiments}\label{section:experiment}
\vspace{-1ex}
\subsection{Experimental Setup}\label{section:model}


To validate the efficacy of our proposed methods in building a controllable Personalized Dialogue (PD) system that addresses the \textit{WWH} problems, we compare the performance of several models. These are enhanced with fine-tuned baseline models, such as dataset blending and negative sampling methods. Additionally, by comparing models trained with different blending weights, we evaluate the impact of the blending weight on the model's grounding propensity and fluency. The baseline models are all derived from our in-house 18B parameter pre-trained language model, which shares the same architecture as GPT-3 \cite{brown2020language}. All experiments are conducted on SKT's proprietary supercomputer, Titan, equipped with NVIDIA A100 SXM4 80GB GPUs.


\begin{table*}[t!]
\centering
\resizebox{\linewidth}{!}{%
\begin{tabular}{lccclccccclcc} 
\toprule
\multirow{3}{*}{Model} & \multicolumn{3}{c}{\textbf{Session / Turn Evaluation}} &  & \multicolumn{8}{c}{\textbf{Grounding Evaluation}} \\ 
\cline{2-4}\cline{6-13}
 & \multirow{2}{*}{Session Score} & \multirow{2}{*}{\begin{tabular}[c]{@{}c@{}}Sensibleness\\(Turn-level)\end{tabular}} & \multirow{2}{*}{\begin{tabular}[c]{@{}c@{}}Specificity\\(Turn-level)\end{tabular}} & \multirow{2}{*}{} & \multicolumn{2}{c}{Hard Grounding} & \multicolumn{2}{c}{Soft Grounding} & \multirow{2}{*}{Sub Total} &  & \multirow{2}{*}{Non-personalized} & \multirow{2}{*}{Total} \\ 
\cline{6-9}
 &  &  &  &  & Consistent & Inconsistent & Consistent & Inconsistent &  &  &  &  \\ 
\cline{1-4}\cline{6-10}\cline{12-13}
$Model_3$ & 0.80 & 0.914 & 0.76 &  & 0.14 (23/162) & N/A (0/0) & 0.23 (9/40) & 1.0 (1/1) & \multicolumn{1}{l}{0.16 (33/203)} &  & 0.03 (10/297) & 0.09 (43/500) \\
$Model_4$ & 0.885 & 0.939 & 0.875 &  & 0.13 (16/125) & N/A (0/0) & 0.17(6/35) & N/A (0/0) & \multicolumn{1}{l}{0.13 (22/160)} &  & 0.03 (11/383) & 0.06 (33/543) \\
\bottomrule
\end{tabular}
}
\vspace{-1ex}
\caption{Results of Subjective Evaluation. 1) Session \& Turn level Evaluation and 2) Grounding Evaluation: the ratio of the count of bad sensible responses to the count of each grounding type described in \ref{section:evaluation}. A bad sensible response means that the response scored a 0 on the "sensible" evaluation.}
\label{tab:subjective_evaluation_result}
\vspace{-2ex}
\end{table*}

\subsection{Evaluation}\label{section:evaluation}
\textbf{Objective Evaluation}
We use perplexity (PPL) to measure the fluency of the responses generated by the model. In addition, the F1 score between the persona attributes and the generated response acts as a proxy to evaluate the model's ability to ground. We also calculate the P-coverage score, which measures how well the user persona is reflected in the generated responses \cite{song2019exploiting}.  \\

\noindent \textbf{Subjective Evaluation}
We complement objective evaluation metrics with subjective human evaluation at both the session and turn levels, specifically employing the Sensibleness and Specificity (SS) score rated as either 0 or 1 at the turn level \cite{adiwardana2020towards}. Particularly, to analyze the pattern and quality of grounded responses at the turn level, we categorize them according to our proposed four grounding types, which are as follows.
First, we assess whether the agent's response, $y$, is personalized. Second, we categorize $y$ based on two criteria: \textbf{grounding level} and \textbf{consistency}.

Under the \textbf{grounding level}, we have two subcategories: 1) \textit{Hard Grounding}, where there's a direct and explicit association between $y$ and the persona attribute, $\rho_\text{pos}$, characterized by high expressive similarity. 2) \textit{Soft Grounding}, where there's an indirect and implicit association between $y$ and $\rho_\text{pos}$, marked by low expressive similarity.

Under the \textbf{consistency} category, we have two subcategories: 1) \textit{Consistent Grounding}, where there's consistency between $y$ and the given $\rho_\text{pos}$. 2) \textit{Inconsistent Grounding}, where there's an inconsistency between $y$ and the given $\rho_\text{pos}$.

\subsection{Results}\label{section:results}

\subsubsection{Effect of Negative Persona Attributes}
Table \ref{tab:objective_evaluation} illustrates the impact of introducing negative persona attributes on persona-grounded response generation. $Model_{1}$ trained with only one given positive persona attribute shows the highest F1 and PPL scores of 11.4 and 0.28, respectively. On the other hand, $Model_{2}$ trained with negative persona attributes has a PPL of 10.97 and an F1 score of 10.15, which is slightly lower than $Model_{1}$. Despite the decrease in grounding frequency, the model demonstrates improved response generation by reasonably selecting an appropriate persona attribute given the dialogue context.
We hypothesize that the PPL increases because the model learns to distinguish the most suitable persona among several persona attributes in a given context. Furthermore, despite the reduced inclination to ground, we observe that the model can still generate high-quality personalized responses at every turn.

\subsubsection{Effect of Negative Persona Subset}
Table \ref{tab:objective_evaluation} illustrates the effectiveness of the negative persona subset in controlling the \textit{WHEN} problem. Through the application of the negative persona subset, $Model_3$ learns to refrain from generating personalized responses when the persona attributes are not appropriate for the given context. In Table \ref{tab:objective_evaluation}, $Model_3$ demonstrates a decrease in persona grounding and a significant increase in fluency compared to $Model_2$, as indicated by the lower PPL, F1, and P-Cover scores (9.37, 01, and 0.05, respectively). We believe the key reason for this enhanced fluency is that the model generates more frequent and natural casual responses to non-personalized turns in the test set, without the need to ground on irrelevant persona subsets.


\subsubsection{Effects of Blending Datasets: Trade-Off between Model Fluency and Grounding}
As shown in Table \ref{tab:blending_weight}, there is a trade-off between the model's fluency and tendency to ground. As the weight of the $\text{MSPD}_\text{NPR}$ dataset with negative persona augmentations increases, the F1 score decreases from 0.14 to 0.06, and the P-cover score falls from 0.06 to 0.1 and 0.05.  
Conversely, the PPL decreases from 10.46 to 9.33. This means that 
an increase in the number of persona augmented negative samples means the model ground less frequently, leading to a more natural conversation flow with better quality responses. 

Achieving natural and engaging conversations requires careful consideration of the trade-off between the model's inclination to ground and response fluency. To control the \textit{WWH} balance, we can adjust the blending weights for datasets with different persona augmentations and select appropriate values for PPL and F1 scores. We set a F1 score of '1' as the minimum threshold for the model's grounding tendency, as we have consistently observed that models with F1 scores below 1 seldom attempt grounding in conversations. This approach ensures that optimal PD systems maintain a balance between a sufficient quantity of grounded responses and a high fluency score.


\subsubsection{Effect of RTL Generation: Enhanced Explainability and Fluency}
As can be seen in Table \ref{tab:objective_evaluation}, based on the F1 score and P-Cover, $Model_4$ trained to generate both RTL and personalized responses, demonstrates little difference in tendency to ground when compared to $Model_3$. On the other hand, we found that the PPL score decreased to 8.88. This result is consistent with \citet{kim2022prosocialdialog}'s research, which showed that the quality of generation was enhanced when information related to the target response was generated simultaneously. 

We also evaluated explainability by analyzing whether the generated Response Type Labels (RTL) accurately reflect the model's decisions on persona grounding. For this purpose, we sampled 90 generated responses for each response type. 
The accuracy of the generated RTL for the casual and the personalized response type wa 96.7\% and 98.8\%, respectively. This confirms that generating the RTL provides a reliable explanation for the model's decision on the \textit{WHEN} problem.

\subsubsection{Subjective Grounding Evaluation}

The high average (over 0.8) scores for both turn and session levels in Table \ref{tab:subjective_evaluation_result} demonstrate that models trained on the high-quality MSPD dataset can generate appropriate responses. In the grounding evaluation, the vast majority of both hard and soft grounding cases demonstrated persona-consistent results. Both models exhibited nearly four times as many hard grounding instances as soft groundings, and they had a lower rate of "bad-sensible" responses. This suggests that the models are strongly inclined to ground persona information in responses in a manner that is both natural and explicit, given the context. Upon closer examination of "bad-sensible" instances of hard grounding, we found that as the models concentrate more on grounding the persona, responses can sometimes become unnatural within the given context. However, the proportion of "bad sensible" grounding responses was in the 10\% range, confirming that the model generally generates high-quality personalized responses.


The RTL generation model ($Model_4$) shows a lower inclination to ground, yet it had a better bad-sensible ratio. Therefore, in accordance with the objective evaluation result, we can confirm that generating both the response and the RTL can have a positive effect on fluency, even though there is no significant improvement in terms of session evaluation.

\subsubsection{Correlation between objective and subjective evaluations}

We confirmed a positive correlation between fluency, as measured by PPL, and human sensibleness judgment. $Model_4$ exhibited a decrease of 0.49 in PPL compared to $Model_3$, indicating improved fluency in Table \ref{tab:objective_evaluation}. In the session and turn evaluation presented in Table 
 \ref{tab:subjective_evaluation_result}, $Model_4$ exhibited significantly higher scores compared to $Model_3$ across all evaluation criteria related to fluency. Turn-level grounding evaluation also revealed a lower bad sensibleness ratio for personalized/non-personalized turns (0.13 and 0.03, respectively), confirming enhanced sensibleness of $Model_4$'s responses. We also found a positive correlation between subjective evaluation (i.e., the amount of grounded generation) and the P-Coverage metric used to assess grounding propensity. In Table \ref{tab:objective_evaluation}, $Model_4$ exhibited a slight decrease in P-Coverage compared to $Model_3$. This corresponds to the reduced number (approximately 40) of personalized turns generated by $Model_4$ in Table \ref{tab:subjective_evaluation_result}, reflecting an actual decrease in the model's grounding propensity. Consequently, considering the cost of subjective evaluation, objective assessment appears feasible for accurately evaluating the model's fluency and grounding tendencies in real-world service operations.

\section*{Conclusion}\label{section:conclusion}
We proposed a method to build a personalized open-domain dialogue system that addresses the \textit{WWH} problem for natural and engaging conversation through weighted dataset blending (\textit{WHEN}), negative persona subsets (\textit{WHEN}), negative persona attributes (\textit{WHAT}), and the creation of highly curated personalized conversation datasets (\textit{HOW}). We also demonstrate that generating a response type label (\textit{RTL}) enhances both the  controllability and explainability of model decisions about the  \textit{WHEN}; this is crucial in commercial service. Experimental results show the effectiveness of our proposed methods in addressing and controlling the \textit{WWH} problem, as seen in both subjective and objective evaluations.

\section*{Acknowledgements}
We would like to express our sincere gratitude to the members of SK Telecom and A.Tech for their dedicated support throughout this project. Special thanks are extended to the members of the Foundation Modeling team for their technical assistance, meaningful discussions, and contributions to improving the model's performance, training, and deployment. Additionally, we would like to thank the linguists from the Dialogue PO team for their invaluable contributions in generating and evaluating high-quality datasets for model training and improvement.




\bibliography{anthology,custom}
\bibliographystyle{acl_natbib}
\clearpage

\appendix

\section{Details of MSPD Dataset}~\label{sec:appendixA}
\subsection{Statistics of MSPD}
\begin{table}[ht]
\label{tab:my-table}
\centering
\resizebox{\linewidth}{!}{%
\begin{tabular}{@{}ll@{}}
\toprule
Types                                        &         \\ \midrule
\# Episodes                                  & 13,469  \\
\# Sessions                                  & 53,880  \\
\# Utterances                                & 601,062 \\
Avg. \# turns per session                    & 11.15   \\
Avg. \# personalized response per session        & 1.90  \\ 
Avg. \# user persona per episode             &  7.18       \\
Avg. \# newly aggregated persona per episode & 2.18    \\
Avg. length of user utterances               &  33.72        \\
Avg. length of agent response                & 28.10        \\ \bottomrule
\end{tabular}
}
\caption{Statistics of the MSPD Dataset}
\end{table} 

\subsection{Model Training Settings}~\label{sec:appendixA_Modeltraining}
For the experiments in our study, we fine-tuned an 18B parameter model with the same architecture as GPT-3 \cite{brown2020language}, but with 40 layers, a hidden size of 6144, and 48 attention heads. The model is trained for a single epoch with a micro batch size of 8, using a learning rate of 1.0e-05. To prevent overfitting, a dropout rate of 0.1 and a weight decay of 1.0e-1 are employed. The input sequence length is 1024. The models in Table \ref{tab:objective_evaluation} 1 are trained with the blending weight set to 0.85 for $D_{casual}$ datasets, 0.7 for $\text{MSPD}_{PR}$, and 0.8 for the $\text{MSPD}_{NPR}$ dataset.

\clearpage
\onecolumn

\section{Examples}~\label{sec:appendixB}
\vspace{-2ex}
\subsection{Example of MSPD Dataset}
\vspace{-2ex}
%

\begin{figure*}[!htbp]
	\centering
	\begin{subfigure}[b]{\textwidth}
		\includegraphics[width=16.5cm, height=23cm]{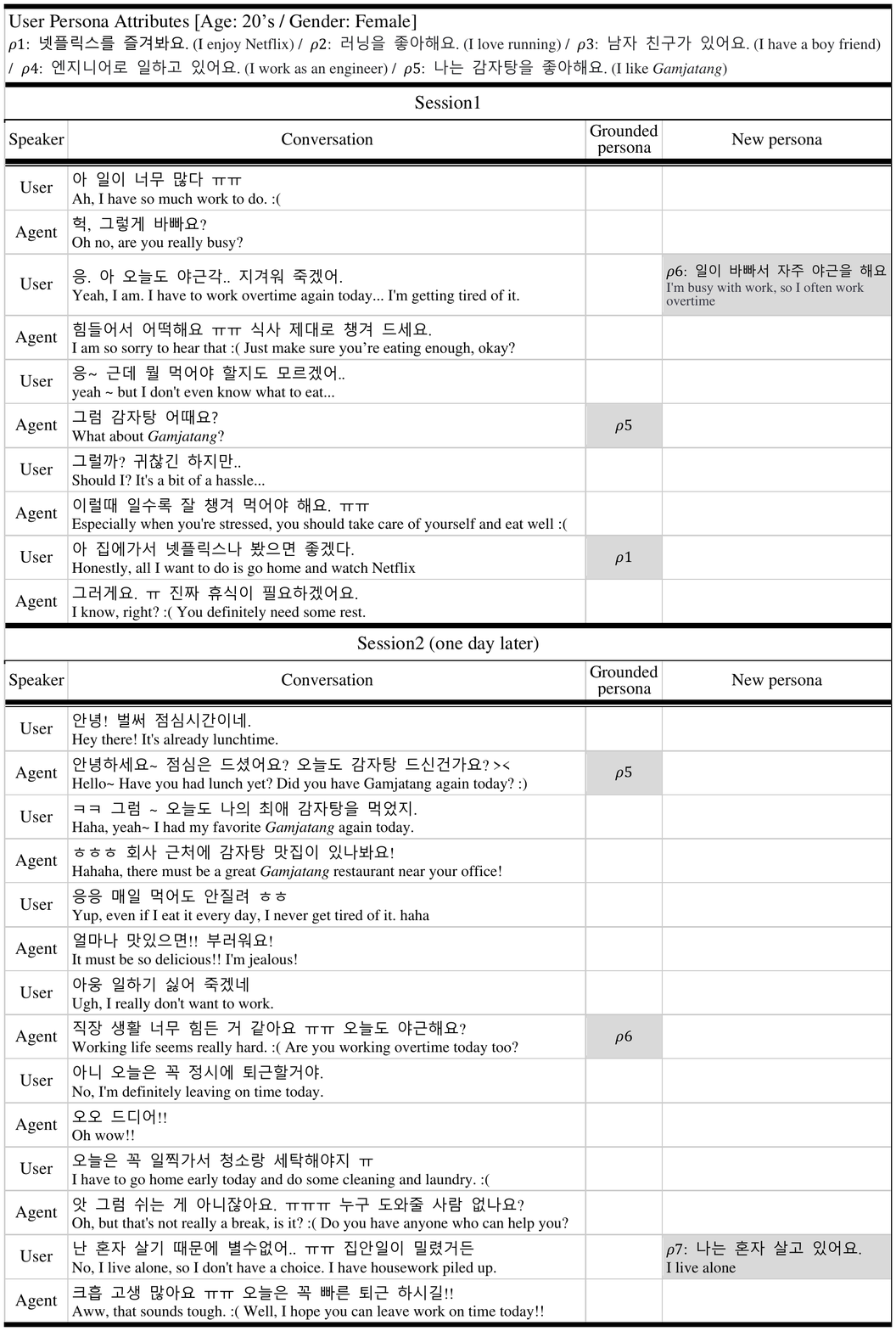}
	\end{subfigure} \
	\vspace*{-10ex}
	\caption{A sample of a multi-session conversation in the MSPD}
	\vspace*{-1ex}
	\label{fig:data_example2}
\end{figure*}

\subsection{Subjective Evaluation}
\begin{figure*}[ht]
    \centering
    \includegraphics[width=\linewidth]{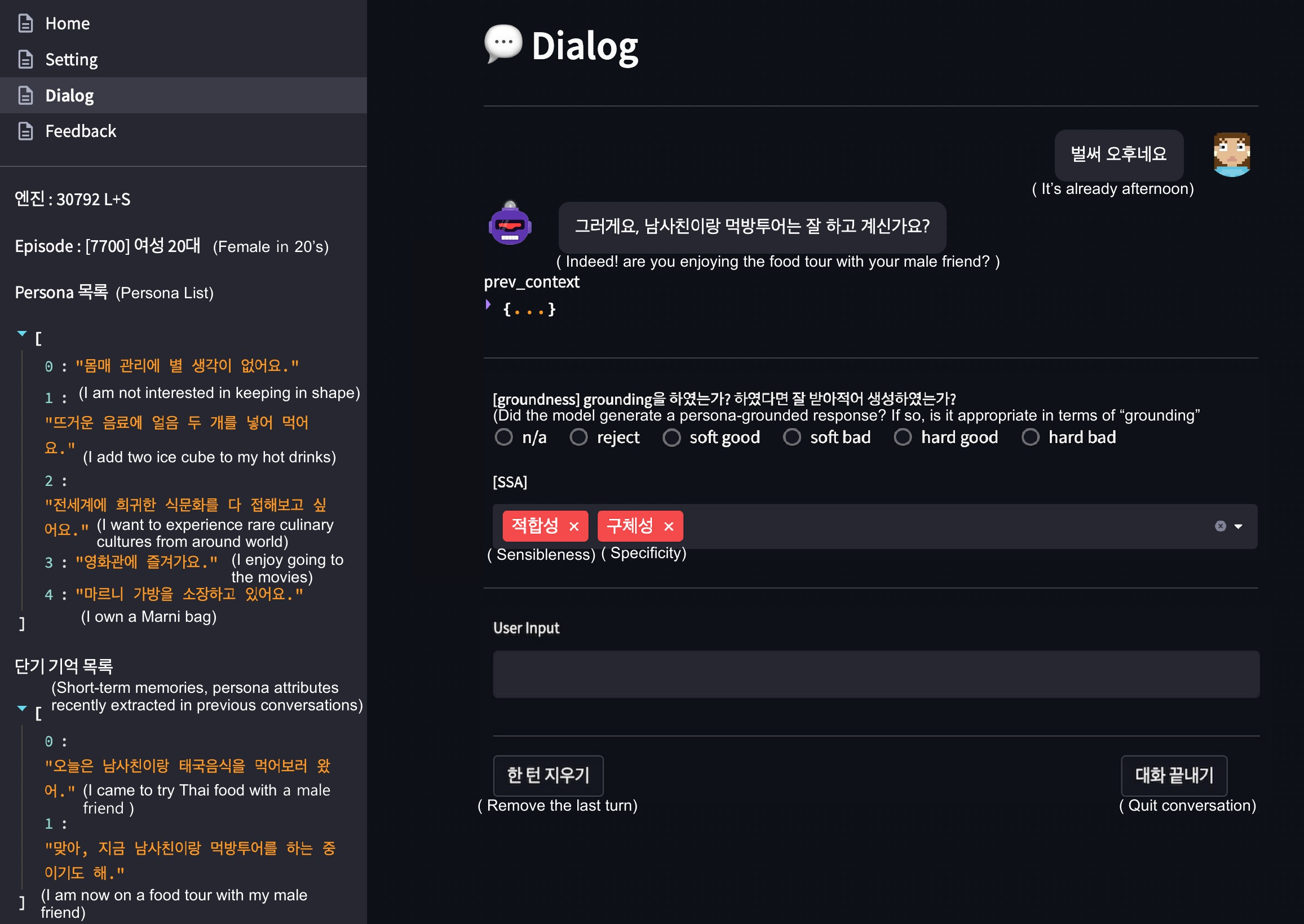}
    \vspace{-3ex}
    \caption{A snapshot of the subjective evaluation Tool. 
    }
    \label{fig:subjective_eval_sample}
\vspace{-2ex}
\end{figure*}

\begin{figure*}[ht]
    \centering
    \includegraphics[width=\linewidth]{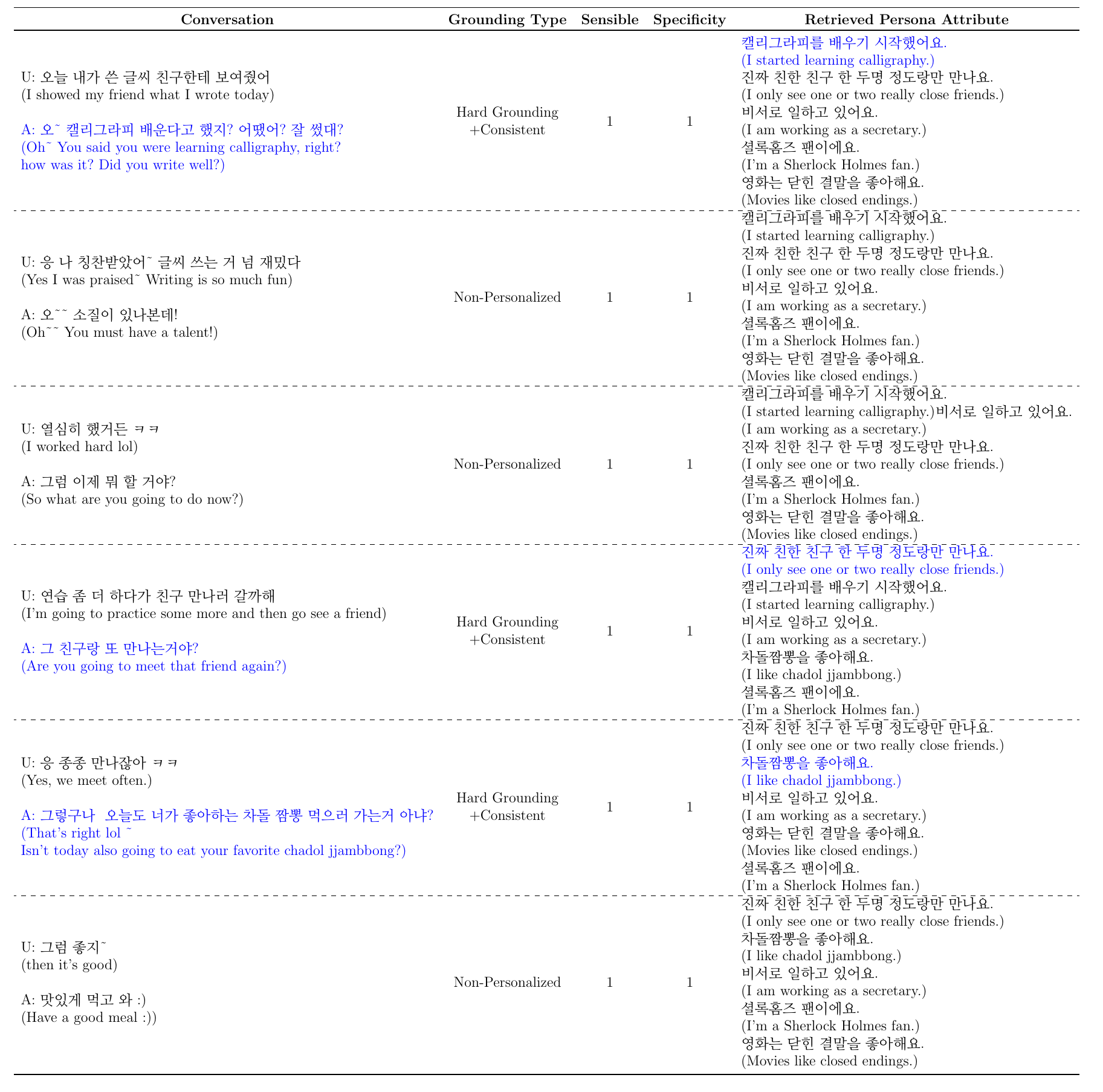}
    \vspace{-4ex}
    \caption{
Example of generations of our personalized dialogue model with subjective evaluation for hard grounding. Blue colored texts are persona-grounded responses and persona attributes on which the model grounds the response}
    \label{fig:subjective_eval_sample2}
\vspace{-1ex}
\end{figure*}

\begin{figure*}[ht]
    \centering
    \includegraphics[width=\linewidth]{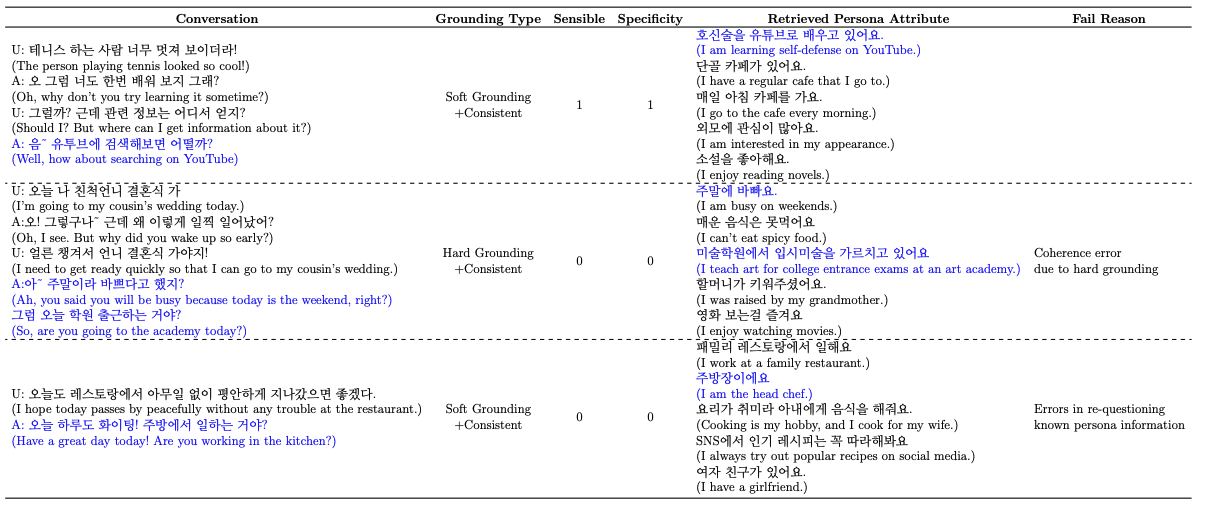}
    \vspace{-3ex}
    \caption{
Example of subjective evaluation for soft grounding and fail cases}
    \label{fig:subjective_eval_sample3}
\vspace{-3ex}
\end{figure*}

\clearpage
\subsection{Deployment Tool: Sanity Testing}
\begin{figure*}[ht]
    \centering
    \includegraphics[width=\linewidth]{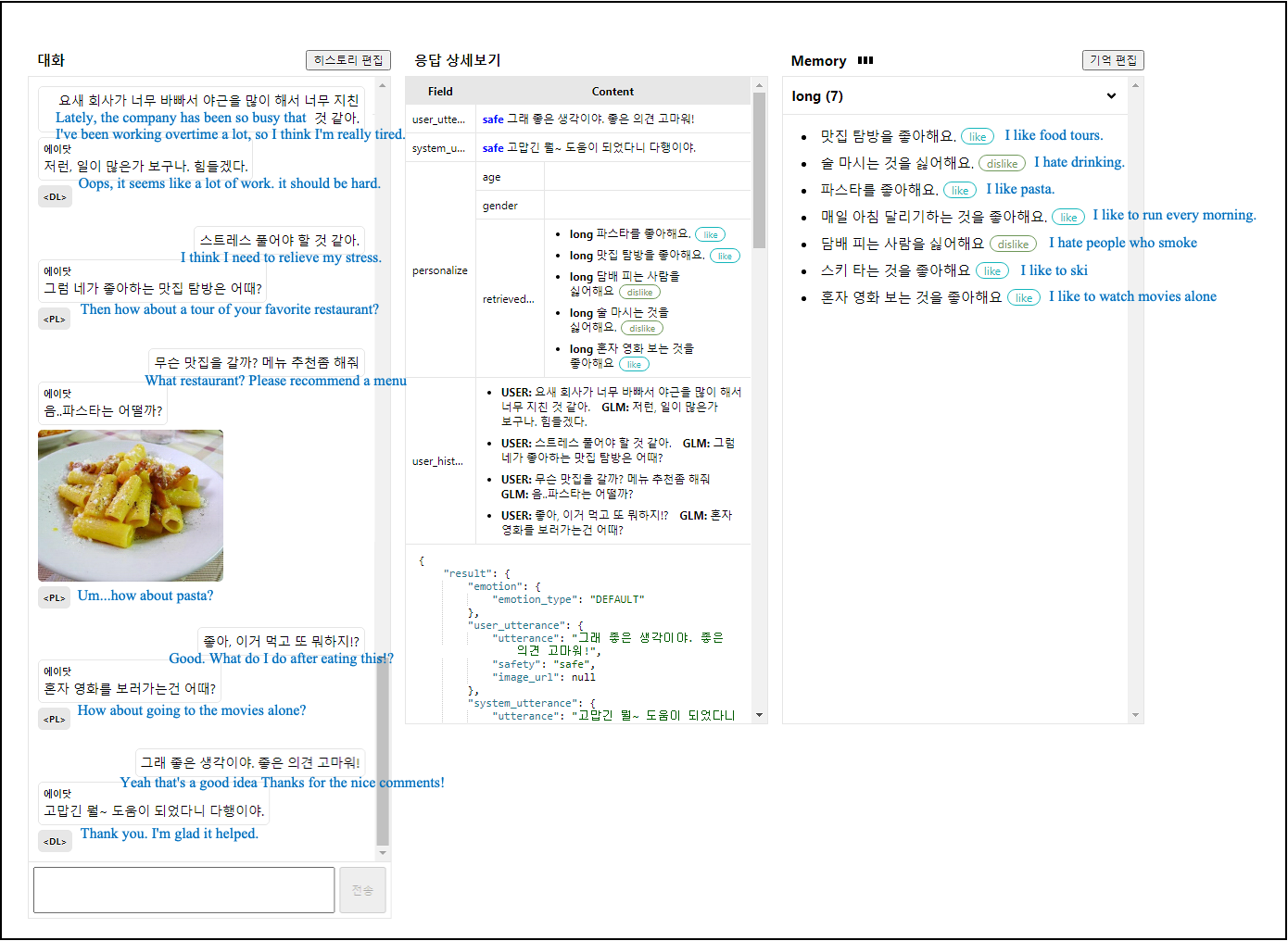}
    \vspace{-3ex}
    \caption{A Snapshot of the Sanity Testing Tool. 1) The leftmost area is for interactive conversation with the agent, and the <PL> and <DL> tags refer to the response type generated by the model; <PL> is a personalized response, and <DL> is a non-personalized response type. 2) The center pane shows information related to the user and the current turn. And 3) the window on the right displays user persona.}
    \label{fig:sanity}
\vspace{-3ex}
\end{figure*}


\end{document}